\documentclass[11pt,a4paper]{article}
\usepackage[hyperref]{acl2021}
\usepackage{times}
\usepackage{tabularx} 
\usepackage{amsmath}
\usepackage{graphicx}
\usepackage{latexsym}

\usepackage{microtype}

\aclfinalcopy 

\title{UniParma at SemEval-2021 Task 5: Toxic Spans Detection Using CharacterBERT and Bag-of-Words Model}

\author{Akbar Karimi \\\And
  Leonardo Rossi \\
  \\
  University of Parma \\
  \texttt{\{akbar.karimi, leonardo.rossi, andrea.prati\}@unipr.it} \\\And
  Andrea Prati \\}

\date{}

\begin{document}
\maketitle
\begin{abstract}
With the ever-increasing availability of digital information, toxic content is also on the rise. Therefore, the detection of this type of language is of paramount importance. We tackle this problem utilizing a combination of a state-of-the-art pre-trained language model (CharacterBERT) and a traditional bag-of-words technique. Since the content is full of toxic words that have not been written according to their dictionary spelling, attendance to individual characters is crucial. Therefore, we use CharacterBERT to extract features based on the word characters. It consists of a CharacterCNN module that learns character embeddings from the context. These are, then, fed into the well-known BERT architecture. The bag-of-words method, on the other hand, further improves upon that by making sure that some frequently used toxic words get labeled accordingly. With a $\sim$4 percent difference from the first team, our system ranked $36^{th}$ in the competition. The code is available for further research and reproduction of the results\footnote{\url{https://github.com/IMPLabUniPr/UniParma-at-semeval-2021-task-5}}.
\end{abstract}

\section{Introduction}
The user generated digital content is increasing rapidly every second of the day. This can include some toxic language whose detection can be difficult due to the complexities of human languages. We address this problem by participating in SemEval Workshop 2021 Task 5 \citep{pav2020semeval}. 

In many cases, the data, which are considered to be toxic, contain words that have not been written in their standard forms. There might also be a lot of misspelling or letter replacements. In addition, usually the words that are considered to be the most offensive are bleeped which makes them difficult to be recognized if we use a model which learns the content representation based on the words. Apart from word related issues, the context also plays a crucial role in the meaning that a word conveys since words in different contexts can have various meanings. 

Therefore, in order to cope with these issues, we opt for a recently pre-trained language model which has been trained on character level. CharacterBERT \citep{el2020characterbert} is a deep neural network model that has been pre-trained on Wikipedia and OpenWebText \citep{Gokaslan2019OpenWeb} corpora using the BERT architecture \citep{devlin2019bert} with an addition of a character-aware Convolutional Neural Network (CNN) \citep{kim2016character, peters2018deep}. BERT-based models have now become pervasive in many different natural language processing tasks such as reading comprehension \citep{xu2019bert}, named entity recognition \citep{liang2020bond}, sentiment analysis \citep{karimi2020improving}, and language understanding \citep{jiao2020tinybert} as well as similar ones to toxic language detection such as propaganda detection \citep{jurkiewicz2020applicaai}. 
While the BERT model is beneficial in extracting the contextual information from the text on the word level, the character-aware CNN attends to the individual letters which helps in dealing with out-of-vocabulary, unknown and rare words. 

In addition to using a deep language model for detecting toxic language, we employ a very simple Bag-of-Words model that can achieve a close performance to that of the deep model. By building a dictionary of toxic words from the training data and by taking into account their frequency and ratio of toxicity, we come up with a simple model that performs as closely as about $2$ percent difference in performance to the deep model's result. Moreover, we improve the results of CharacterBERT by combining it with the output of a version of the Bag-of-Words model. 

\begin{table*}[hbt!]
    \centering
    \begin{tabularx}{\textwidth}{X|X}
        \textbf{spans} & \textbf{text} \\
        \hline
        [8, 9, 10, 11, 12, 13, 14, 15, 16, 17, 18, 19, 20, 21, 22, 23, 24, 25, 26, 27, 28, 29, 30, 31, 32, 33, 34, 35, 36, 37, 38, 39] & Another violent and aggressive immigrant killing a innocent and intelligent US Citizen.... Sarcasm \\
        \hline
       [0, 1, 2, 3] & Damn, a whole family. Sad indeed. \\
        \hline
       [7, 8, 9, 10, 11, 12, 13, 14, 15, 16, 17] & What a knucklehead. How can anyone not know this would be offensive?? \\
       \hline
    \end{tabularx}
    \caption{Three examples from the training set}
    \label{tab:examples}
\end{table*}

\section{System Description}

Our system consists of four main stages namely, pre-processing, applying CharacterBERT, applying Bag-of-Words model, and finally combining the results of the two models. We describe each of these stages in the following subsections.

\subsection{Pre-processing}

The training dataset consists of rows of various lengths and an array of character spans indicating their toxic parts. Each row can contain several sentences. Table \ref{tab:examples} shows three examples of the training data.

We approach the task of toxic spans detection as a sequence labelling task where each word of the input row is classified into one of the predefined classes. We define three classes of \{B, I, O\}, meaning that each word can be the first word (B) of a set of continuous toxic words, in between (I), or not toxic (O). Therefore, in order for both of our models to be able to process these inputs, we first need to break the rows into words and label them as one of the above-mentioned classes. This was carried out by simply splitting each input row at the $space$ characters. Then, after creating a dataset that has been labeled on the word level, we can use it as the input of our models. 
The same is done for the Bag-of-Words model with a difference in treating the bleeped words which is described in Subsection \ref{bow-description}. 

\subsection{CharacterBERT}
\begin{figure}
    \centering
    \includegraphics[scale=0.35]{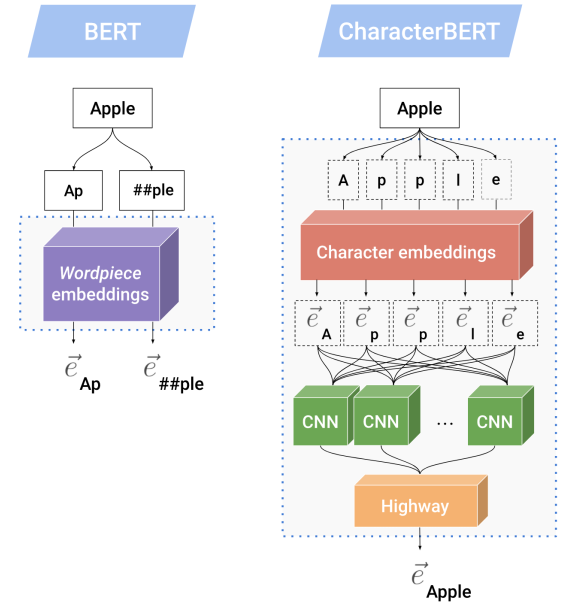}
    \caption{The difference between the BERT and CharacterBERT models is the way they compute the initial embeddings. The former uses word-piece embeddings while the latter uses character embeddings. Figure taken from \citet{el2020characterbert}.}
    \label{fig:charBERT}
\end{figure}

CharacterBERT model is almost identical to the well-known BERT model with a difference in initial embedding. In the general BERT model, words are broken into pieces and the embeddings for these word pieces are computed. In CharacterBERT, however, words are divided into letters or characters. Then, using CNN modules the embeddings are computed on the character level (Figure \ref{fig:charBERT}). This makes the network extract features on the lowest level, making it suitable for contexts which contain many unseen vocabulary terms such as misspelled words or technical jargon. After the initial character-aware CNN layer, there is the BERT\textsubscript{base} architecture which contains $12$ layers (blocks) of Transformer \citep{vaswani2017attention} with the hidden size of $768$ and $12$ attention heads. The final layer representations are converted into logits using a fully connected layer after which a Softmax layer is applied to extract the token's (word's) class.

\begin{figure*}[t]
    \centering
    \includegraphics[scale=0.13]{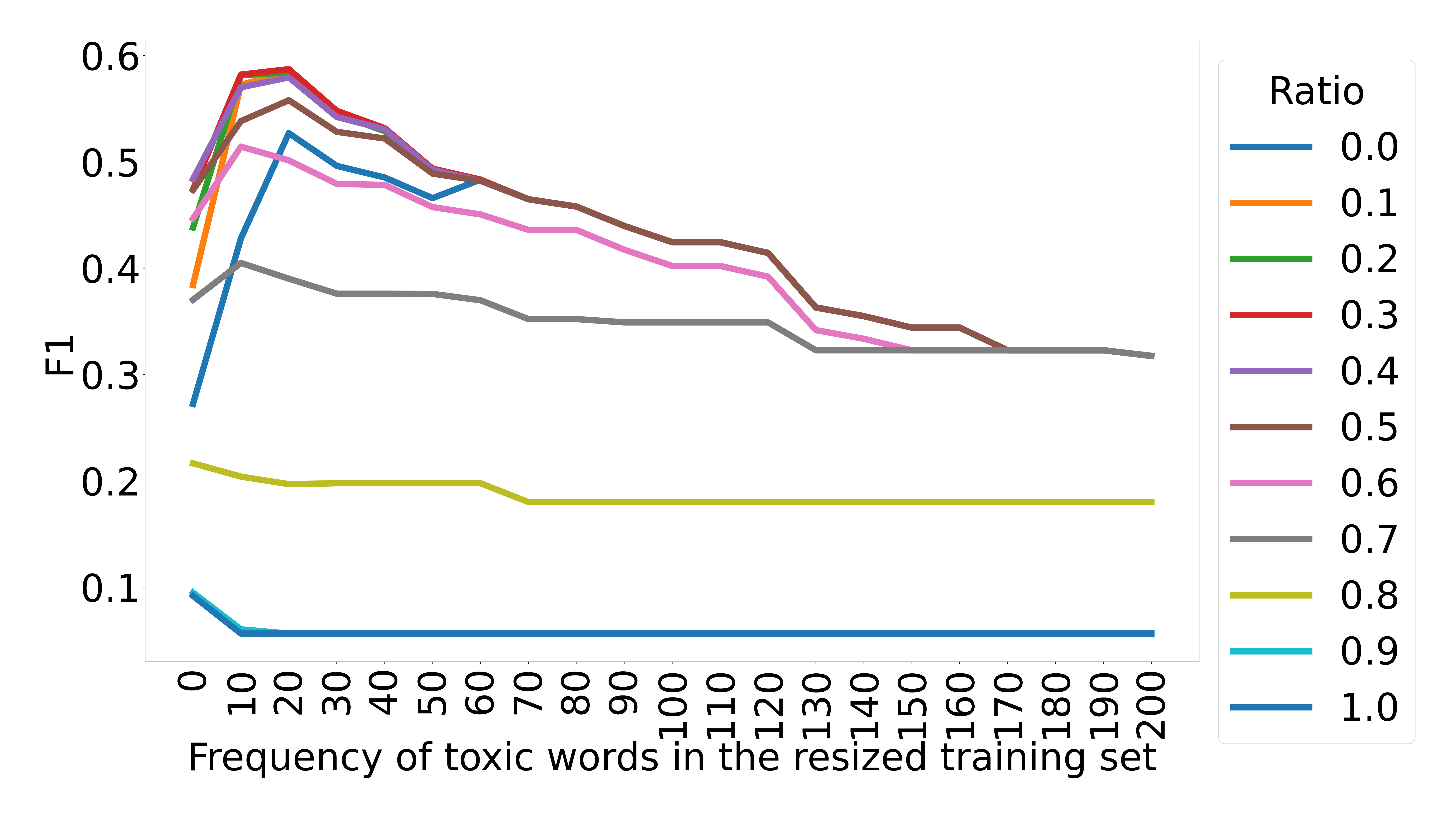}
    \caption{Performance of the Bag-of-Words model on validation set. The frequencies and ratios are the minimum thresholds that specify whether or not to consider a word toxic.}
    \label{fig:bagofwords}
\end{figure*}

\subsection{Bag-of-Words Model}\label{bow-description}
This model is a simple script of fewer than 80 lines of code. However, its performance on the Toxic Spans Detection task can get very close to the CharacterBERT model which has millions of parameters. In this model, by examining the training set, we first build a dictionary of toxic words with their frequency. Table \ref{tab:toxic_words} presents the top ten words of this dictionary in terms of frequency. 

\begin{table}[]
    \centering
    \begin{tabular}{l|c}
        \textbf{Word} & \textbf{Frequency} \\
        \hline
        stupid & 973 \\
        idiot & 557 \\
        idiots & 353 \\
        stupidity & 223 \\
        ignorant & 190 \\
        dumb & 157 \\
        moron & 147 \\
        fool & 141 \\
        pathetic & 138 \\
        crap & 121 
    \end{tabular}
    \caption{Top 10 toxic words in the training set}
    \label{tab:toxic_words}
\end{table}

Then, we locate the words from the toxic dictionary in each sentence of the test set. If the word is found and its frequency as well as its toxicity ratio in the training set are higher than certain values, it is labeled as toxic. This ratio which we call \textit{toxicity ratio} (defined below) along with the \textit{term frequency} are the only parameters of the Bag-of-Words model.
\[
 \textbf{toxicity ratio} = \frac{\text{labeled as toxic frequency}}{\text{total frequency}}
\]

The test dataset also contains words that are bleeped. Since these words can be considered toxic with a high certainty (otherwise they would not be bleeped), we extract them separately from the test set and label them directly as toxic. 

\subsection{Combining the Two Models}

In order to get the improved version of the toxic language labeling, the union of the spans detected by the bag-of-words model and that of CharacterBERT is taken. The results will improve if there are words labeled correctly with the Bag-of-Words model that are not in the output for CharacterBERT. This can be achieved by specifying a high toxicity ratio for a word to be labeled as toxic. Also, the wrongly labeled tokens should not be too many since it can have a negative effect on the F1 score. Therefore, the frequency with which a toxic word appears should be somewhat high. Striking a balance between these two parameters can help improve the output of CharacterBERT. 

\section{Experiments and Results}

\subsection{Performance of CharacterBERT} We ran the experiments for the general domain CharacterBERT with its default setting on a GPU (GeForce RTX 2070) which had 8GB of memory. We specified batch sizes of $4$ for both training and testing and fine-tuned it on the toxic data only for one epoch which produces an F1 score of $65.13$. It is worth noting that more training did not improve the performance. 

\subsection{Analysis of the Bag-of-Words model} 

In order to experiment with the Bag-of-Words model, we divide the original training set into a resized training set with $7000$ sentences and a validation set with $939$ sentences which were taken from the end of the original training set. Then, we find the best parameters on the validation set and using those parameters on the test data, we get a performance of almost $63$ percent which is only $2$ percent smaller than our deep model. Figure \ref{fig:bagofwords} shows this model's performance for its two parameters on the validation set. One parameter represents the \textit{minimum} frequency with which a toxic word appears in the resized training set and the other one is its \textit{minimum} toxicity ratio in the resized training data. 

We can see from Figure \ref{fig:bagofwords} that the best results are achieved when the minimum frequency is $20$ and the minimum ratio is $0.3$ or $0.4$. Since a larger ratio can be a sign of more toxicity, we choose $0.4$ as the ratio and a frequency of $20$ as the thresholds with which we apply the model on the test set. This gives an F1 score of $62.79$ percent (Table \ref{tab:results}) which is not that much below the result of the deep model.

We can also see from Table \ref{tab:results} that although combining the output of the Bag-of-Words model with that of CharacterBERT improves the results a little bit, it is still is not as significant as the first version. In the first version of the Bag-of-Words model, which was found during our primary experiments, the minimum word frequency is $40$ and the minimum toxicity ratio is $0.7$. With these parameters, only 10 words are selected from the training set. The frequency and toxicity ratio of these words can be seen in Table \ref{tab:words}. 

\begin{table}
    \centering
    \begin{tabular}{l|c}
    \hline
        \textbf{Model} & \textbf{F1} \\
        \hline
        CharacterBERT & 65.13 \\
        BoW (v1) & 51.75 \\
        BoW with best parameters (v2) & 62.79 \\
        CharacterBERT + BoW (v2) & 65.87 \\
        \hline
        \textbf{CharacterBERT + BoW (v1)} & \textbf{66.72} \\
        \hline
    \end{tabular}
    \caption{Comparing results of the proposed models. The boldfaced one is the submitted version. BoW: Bag-of-Words model.}
    \label{tab:results}
\end{table}

\begin{table}
    \centering
    \begin{tabular}{l|c|c}
         \textbf{Word} & \textbf{Frequency} & \textbf{Toxicity Ratio} \\
         \hline
         stupid & 973 & 0.78 \\
         idiot & 557 & 0.84 \\
         idiots & 353 & 0.81 \\
         stupidity & 223 & 0.77 \\
         moron & 147 & 0.71 \\
         idiotic & 98 & 0.74 \\
         hypocrite & 75 & 0.88 \\
         shit & 56 & 0.72 \\
         scum & 52 & 0.70 \\
         hypocrites & 44 & 0.76 \\
    \end{tabular}
    \caption{Words selected as the toxic words with minimum frequency of 40 and minimum toxicity ratio of 0.7 (BoW (v1))}
    \label{tab:words}
\end{table}

Although the performance of this version is a lot lower than the second version (v2) of the BoW model, it helps to improve the performance of CharacterBERT. The reason for this behavior can be attributed to the fact that models with higher thresholds both in terms of frequency and toxicity ratio tend to output more certain results, albeit fewer words than the ones that should be labeled as toxic. Therefore, many toxic words that are less probable are not extracted and F1 score drops. 

\begin{figure}
    \centering
    \includegraphics[scale=0.57]{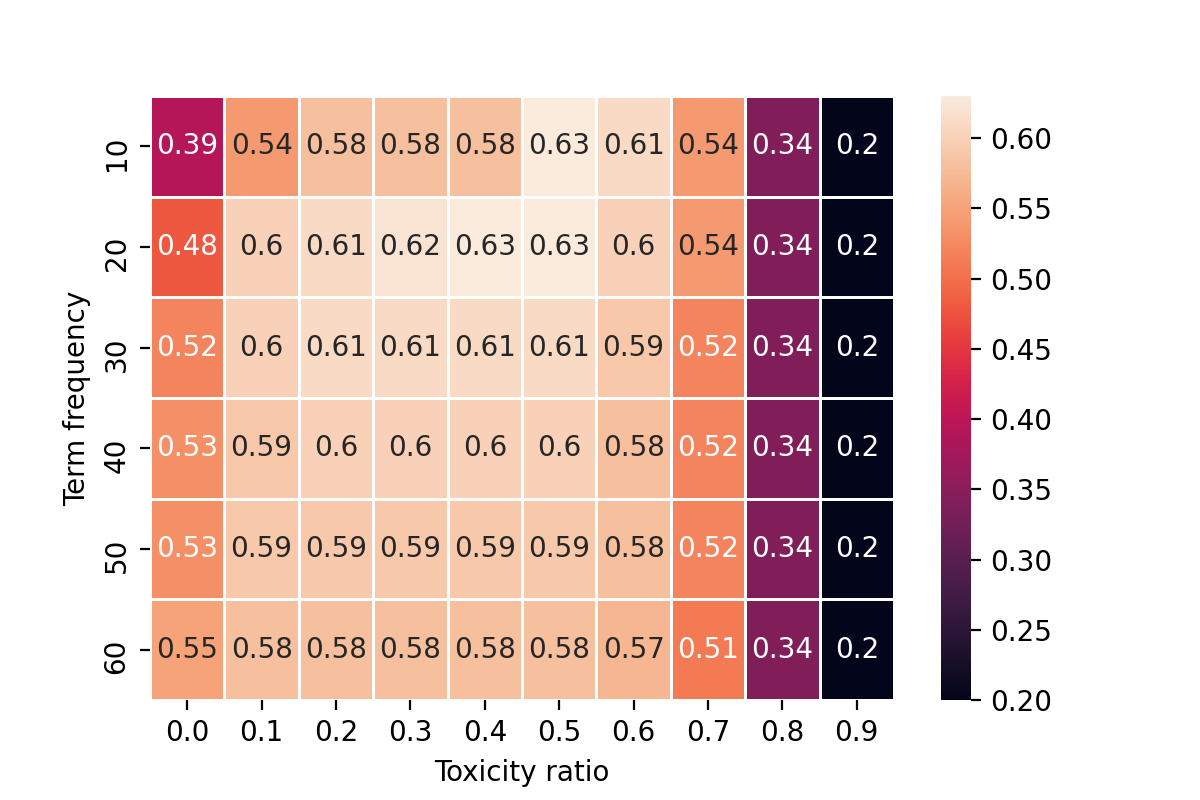}
    \caption{Heatmap of the results (F1 scores) with different values of term frequency and toxicity ratio before combining with CharacterBERT}
    \label{heatmapone}
\end{figure}

\begin{figure}
    \centering
    \includegraphics[scale=0.57]{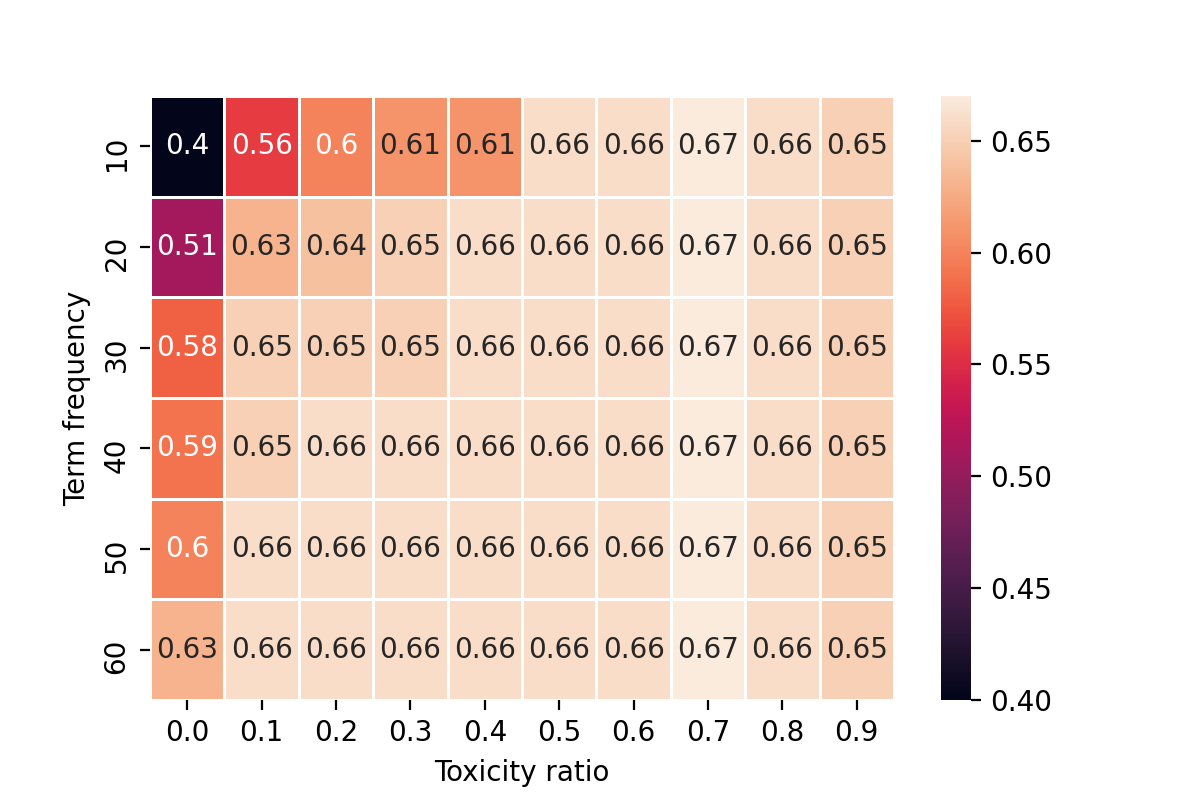}
    \caption{Heatmap of the results (F1 scores) with different values of term frequency and toxicity ratio after combining with CharacterBERT}
    \label{heatmaptwo}
\end{figure}

Looking at Figure \ref{heatmapone}, we can see that, indeed, the best parameters from the experiments on the validation set (ratios $0.3$ and $0.4$ with frequencies $10$ and $20$) yield some of the best results on the test set. However, when these results are combined with the output of the CharacterBERT, we see that the higher the toxicity ratio the better the results (Figure \ref{heatmaptwo}) until $0.7$ which gives the maximum improvement. The $0.8$ ratio makes the predictions still a little better but $0.9$ does not affect them since the words that are labeled as toxic with this certainty have most probably been found also by CharacterBERT.

\section{Conclusion}

We described the system we utilized to detect toxic language. In our approach, we first fine-tune CharacterBERT, a character-level pre-trained language model, on the toxic training data. Then using a simple bag-of-words model, we further improve the results of this system. The Bag-of-Words model labels the words based on their frequency and the ratio of toxicity in the training data. We showed that this model, although extremely simple, gives a close performance to that of CharacterBERT with millions of parameters.
\bibliographystyle{acl_natbib}
\bibliography{anthology,acl2021}

\begin{thebibliography}{12}
\expandafter\ifx\csname natexlab\endcsname\relax\def\natexlab#1{#1}\fi

\bibitem[{Devlin et~al.(2019)Devlin, Chang, Lee, and
  Toutanova}]{devlin2019bert}
Jacob Devlin, Ming-Wei Chang, Kenton Lee, and Kristina Toutanova. 2019.
\newblock {BERT}: Pre-training of deep bidirectional transformers for language
  understanding.
\newblock In \emph{Proceedings of the 2019 Conference of the North American
  Chapter of the Association for Computational Linguistics: Human Language
  Technologies, Volume 1 (Long and Short Papers)}, pages 4171--4186.

\bibitem[{El~Boukkouri et~al.(2020)El~Boukkouri, Ferret, Lavergne, Noji,
  Zweigenbaum, and Tsujii}]{el2020characterbert}
Hicham El~Boukkouri, Olivier Ferret, Thomas Lavergne, Hiroshi Noji, Pierre
  Zweigenbaum, and Jun’ichi Tsujii. 2020.
\newblock {CharacterBERT}: Reconciling {ELMo} and {BERT} for word-level
  open-vocabulary representations from characters.
\newblock In \emph{Proceedings of the 28th International Conference on
  Computational Linguistics}, pages 6903--6915.

\bibitem[{Gokaslan and Cohen()}]{Gokaslan2019OpenWeb}
Aaron Gokaslan and Vanya Cohen.
\newblock Openwebtext corpus.

\bibitem[{Jiao et~al.(2020)Jiao, Yin, Shang, Jiang, Chen, Li, Wang, and
  Liu}]{jiao2020tinybert}
Xiaoqi Jiao, Yichun Yin, Lifeng Shang, Xin Jiang, Xiao Chen, Linlin Li, Fang
  Wang, and Qun Liu. 2020.
\newblock {TinyBERT}: Distilling {BERT} for natural language understanding.
\newblock In \emph{Proceedings of the 2020 Conference on Empirical Methods in
  Natural Language Processing: Findings}, pages 4163--4174.

\bibitem[{Jurkiewicz et~al.(2020)Jurkiewicz, Borchmann, Kosmala, and
  Gralinski}]{jurkiewicz2020applicaai}
Dawid Jurkiewicz, {\L}ukasz Borchmann, Izabela Kosmala, and Filip Gralinski.
  2020.
\newblock {ApplicaAI} at semeval-2020 task 11: On {RoBERTa-CRF}, span {CLS} and
  whether self-training helps them.
\newblock In \emph{Proceedings of the Fourteenth Workshop on Semantic
  Evaluation}, pages 1415--1424.

\bibitem[{Karimi et~al.(2020)Karimi, Rossi, and Prati}]{karimi2020improving}
Akbar Karimi, Leonardo Rossi, and Andrea Prati. 2020.
\newblock Improving {BERT} performance for aspect-based sentiment analysis.
\newblock \emph{arXiv preprint arXiv:2010.11731}.

\bibitem[{Kim et~al.(2016)Kim, Jernite, Sontag, and Rush}]{kim2016character}
Yoon Kim, Yacine Jernite, David Sontag, and Alexander Rush. 2016.
\newblock Character-aware neural language models.
\newblock In \emph{Proceedings of the AAAI conference on artificial
  intelligence}, volume~30.

\bibitem[{Liang et~al.(2020)Liang, Yu, Jiang, Er, Wang, Zhao, and
  Zhang}]{liang2020bond}
Chen Liang, Yue Yu, Haoming Jiang, Siawpeng Er, Ruijia Wang, Tuo Zhao, and Chao
  Zhang. 2020.
\newblock Bond: Bert-assisted open-domain named entity recognition with distant
  supervision.
\newblock In \emph{Proceedings of the 26th ACM SIGKDD International Conference
  on Knowledge Discovery \& Data Mining}, pages 1054--1064.

\bibitem[{Pavlopoulos et~al.(2021)Pavlopoulos, Laugier, Sorensen, and
  Androutsopoulos}]{pav2020semeval}
John Pavlopoulos, Léo Laugier, Jeffrey Sorensen, and Ion Androutsopoulos.
  2021.
\newblock Semeval-2021 task 5: Toxic spans detection (to appear).
\newblock In \emph{Proceedings of the 15th International Workshop on Semantic
  Evaluation}.

\bibitem[{Peters et~al.(2018)Peters, Neumann, Iyyer, Gardner, Clark, Lee, and
  Zettlemoyer}]{peters2018deep}
Matthew Peters, Mark Neumann, Mohit Iyyer, Matt Gardner, Christopher Clark,
  Kenton Lee, and Luke Zettlemoyer. 2018.
\newblock Deep contextualized word representations.
\newblock In \emph{Proceedings of the 2018 Conference of the North American
  Chapter of the Association for Computational Linguistics: Human Language
  Technologies, Volume 1 (Long Papers)}, pages 2227--2237.

\bibitem[{Vaswani et~al.(2017)Vaswani, Shazeer, Parmar, Uszkoreit, Jones,
  Gomez, Kaiser, and Polosukhin}]{vaswani2017attention}
Ashish Vaswani, Noam Shazeer, Niki Parmar, Jakob Uszkoreit, Llion Jones,
  Aidan~N Gomez, {\L}ukasz Kaiser, and Illia Polosukhin. 2017.
\newblock Attention is all you need.
\newblock In \emph{Proceedings of the 31st International Conference on Neural
  Information Processing Systems}, pages 6000--6010.

\bibitem[{Xu et~al.(2019)Xu, Liu, Shu, and Philip}]{xu2019bert}
Hu~Xu, Bing Liu, Lei Shu, and S~Yu Philip. 2019.
\newblock {BERT} post-training for review reading comprehension and
  aspect-based sentiment analysis.
\newblock In \emph{Proceedings of the 2019 Conference of the North American
  Chapter of the Association for Computational Linguistics: Human Language
  Technologies, Volume 1 (Long and Short Papers)}, pages 2324--2335.

\end{thebibliography}


\end{document}